\documentclass{article}
\usepackage{ijcai21}

\usepackage{times}

\usepackage{soul}
\usepackage{url}
\usepackage[hidelinks]{hyperref}
\usepackage[utf8]{inputenc}
\usepackage[small]{caption}
\usepackage{graphicx}
\usepackage{amsmath}
\usepackage{booktabs}
\usepackage{subcaption}
\usepackage{algorithm}
\usepackage{algpseudocode}
\DeclareMathOperator*{\argmax}{arg\,max}
\DeclareMathOperator*{\argmin}{arg\,min}

\title{Efficient Training of Robust Decision Trees Against Adversarial Examples}

\author{
Dani\"el Vos \and Sicco Verwer
\affiliations
Delft University of Technology
\emails
D.A.Vos@tudelft.nl, S.E.Verwer@tudelft.nl
}

\begin{document}

\maketitle

\begin{abstract}
In the present day we use machine learning for sensitive tasks that require models to be both understandable and robust. Although traditional models such as decision trees are understandable, they suffer from adversarial attacks. When a decision tree is used to differentiate between a user's benign and malicious behavior, an adversarial attack allows the user to effectively evade the model by perturbing the inputs the model receives. We can use algorithms that take adversarial attacks into account to fit trees that are more robust. In this work we propose an algorithm, \textsc{groot}, that is two orders of magnitude faster than the state-of-the-art-work while scoring competitively on accuracy against adversaries. \textsc{groot} accepts an intuitive and permissible threat model. Where previous threat models were limited to distance norms, we allow each feature to be perturbed with a user-specified parameter: either a maximum distance or constraints on the direction of perturbation. Previous works assumed that both benign and malicious users attempt model evasion but we allow the user to select which classes perform adversarial attacks. Additionally, we introduce a hyperparameter $\rho$ that allows \textsc{groot} to trade off performance in the regular and adversarial settings.
\end{abstract}

\section{Introduction} \label{intro}
Machine learning algorithms have been applied to a wide range of sensitive tasks in areas such as cyber security, health care and finance \cite{buczak_survey_2015,chen_disease_2017}. In these tasks it is important that decisions made by the model are explainable \cite{samek_explainable_2017} and that the model is robust against adversarial attacks. When we want to train a model to detect credit card fraud for example, we want to both understand how the model predicts fraud and we want to make it difficult for users to evade the model. Recently it has been shown that neural networks \cite{szegedy_intriguing_2014,goodfellow_adversarial_2014} and similarly linear models, decision trees and support vector machines \cite{papernot_transferability_2016} are vulnerable to such evasions called adversarial examples: perturbed samples that trick the model into misclassifying them.

Many solutions have been proposed to increase the adversarial robustness of neural networks \cite{papernot_distillation_2016,meng_magnet_2017,samangouei_defense_gan_2018,ilyas_adversarial_2019} but these models are inherently hard to explain. These solutions can also apply to linear classifiers, but such models perform poorly on data that is not linearly separable. Decision trees are a popular choice of explainable model and recently the first steps have been taken to train them robustly:

Kantchelian et al. \cite{kantchelian_evasion_2016} proposed a hardening approach for decision tree ensembles using a mixed-integer linear programming approach and introduced an optimal attack. Chen et al. \cite{chen_robust_2019} trained robust trees by scoring with the worst-case information gain or Gini impurity criteria then proposed a fast approximation. Calzavara et al. \cite{calzavara_treant_2019} created TREANT, a flexible approach in which the user describes a threat model using axis-aligned rules and a loss function under attack that it directly optimizes.

In this work we propose \textsc{groot}, a new algorithm for training robust decision trees. Like Chen et al. \cite{chen_robust_2019}, we closely mimic the greedy recursive splitting strategy that traditional decision trees use and we score splits with the worst-case Gini impurity. Additionally, we propagate perturbed samples to child nodes and search through more candidate splits. Since the adversarial Gini impurity is concave with respect to the number of modified data points, we can compute its value in constant time using an analytical solution. This makes \textsc{groot} two orders of magnitude faster than TREANT \cite{calzavara_treant_2019}. Moreover, it scores competitively on adversarial accuracy while accepting an intuitive and permissible threat model. 
Where previous works \cite{kantchelian_evasion_2016,chen_robust_2019,calzavara_treant_2019} focus on robustness against all attacks, we recognize that models might be attacked in just a fraction of cases. We encode this in a parameter $\rho$ that trades off accuracy against regular users and robustness against attackers without influencing run-time. Our main contributions are:

\begin{itemize}
    \item An efficient score function that allows us to fit trees two orders of magnitude faster than the state of the art.
    \item An algorithm that achieves competitive performance to the state of the art in the adversarial setting.
    \item A flexible implementation that allows users to specify attacks in terms of axis aligned perturbations.
\end{itemize}

\section{Background and Related Work}
\label{sec:background}

\subsection{Decision Tree Learning}
A well-researched and interpretable model is the decision tree. Decision trees are comprised of nodes that perform a decision on a single feature of the dataset, sending some samples to a left child and others to the right. When we follow a path through these nodes we eventually reach a leaf that contains a prediction value. In this work we focus on two-class decision trees containing both decision nodes on numerical and categorical variables. Decision trees have recently gained more interest after their success in ensembles such as random forests and gradient boosting. 
Particularly approaches such as XGBoost \cite{chen_xgboost_2016} have dominated in recent machine learning competitions. It is worth noting that when used in large ensembles, the models lose their interpretability as the number of decision regions grows exponentially with the number of trees.

Popular methods for fitting decision trees greedily choose the best split according to the Gini impurity (CART \cite{breiman_cart_1984}) or information gain (ID3 \cite{quinlan_id3_1986}) criterion, then recursively continue this operation. This work focuses on the Gini impurity as the two criteria differ in their behavior only in $2\%$ of the cases \cite{raileanu_theoretical_2004}, but the information gain criterion requires an expensive logarithm operation to compute.

\subsection{Adversarial Examples}
When machine learning models are deployed in the real world, one can expect them to be attacked. One such attack is evasion, where a user will submit an input with specifically crafted perturbations so that it is misclassified. These perturbed inputs are known as `adversarial examples' \cite{szegedy_intriguing_2014}. Generally, one assumes that an attacker can move samples within some limits surrounding the samples, e.g. a region of radius $\epsilon$ in the $L_\infty$ norm. In this work we allow users to specify these regions with a flexible specification and assume that the attacker knows everything about the model. This constitutes a white-box scenario as opposed to black-box in which the attacker can only use the model as an oracle.

\subsection{Related Work}
We briefly summarize three previous works in the area of adversarial learning decision trees. In Table \ref{algorithms-overview} we compare each algorithm's time complexity and threat model.

\begin{table}
    \centering
    \setlength{\tabcolsep}{4pt}
    \caption{Overview of algorithms for fitting robust decision trees. Where $n$ is the number of samples, $f$ number of features, $d$ tree depth, $r$ number of axis-aligned rules, $X$ solver time and $i$ hardening iterations. Recent works move away from $L$ norms to support more flexible specifications.}
    \begin{tabular}{@{}lll@{}}
        \toprule
        Algorithm & Time complexity & Threat model\\
        \midrule
        \textsc{groot} (ours) & $\mathcal{O}(n\log{n} \: 2^d \: f)$ & $L_{\infty}$ and variations\\
        TREANT & $\mathcal{O}(n^2 \: 2^d \: 2^r \: f)$ & axis-aligned rules\\
        Robust trees & $\mathcal{O}(n^2 \: 2^d \: f)$ & $L_{\infty}$\\
        MILP hardening & $\mathcal{O}(n\log{n} \: X \: 2^d \: f)$ & $L_0$ / $L_1$ / $L_2$ / $L_{\infty}$\\ 
        Approx. hardening & $\mathcal{O}(n\log{n} \: i \: d2^d \: f)$ & $L_0$\\
        Regular trees & $\mathcal{O}(n\log{n} \: 2^d \: f)$ & $-$\\
        \bottomrule
    \end{tabular}
    \label{algorithms-overview}
\end{table}

\subsubsection{Hardening Tree Ensembles}
Setting the foundations of robust decision trees, Kantchelian et al. \cite{kantchelian_evasion_2016} propose a hardening approach for tree ensembles. They show that finding adversarial examples under distance constraints is NP-hard for tree ensembles and they provide a MILP translation of the problem for arbitrary $L_x$ norm. Using MILP for hardening is complete but inefficient, they therefore also give an approximation algorithm for attacking a model under the $L_0$ norm constraints. Although hardening using the approximate attack shows an increase in classifier robustness under $L_0$ norm attacks, the classifier performs worse under different norms. This highlights that the threat model during training needs to closely represent the threat model at test time. The hardening approach can not be used to fit a single robust tree.

\subsubsection{Robust Decision Trees Against $L_\infty$ Attacks}
Chen et al. \cite{chen_robust_2019} assume an attacker that is restricted by an $L_\infty$ norm and give an algorithm to fit a tree that is robust to these attacks by using the `information gain under attack'. This criterion is the information gain for a situation in which the attacker moves points within an $L_\infty$ radius to confuse the classifier maximally. In general terms Chen et al.'s algorithm finds the best split, moves points near the split over it if that reduces the model's information gain and recursively continues this splitting procedure on the left and right side of the split. This algorithm closely follows the greedy tree building approach for normal decision trees and intuitively increases robustness against an $L_\infty$ threat model.

\subsubsection{TREANT}
TREANT \cite{calzavara_treant_2019} introduces a more flexible approach to specifying attacker capabilities. By allowing the user to describe an adversary using axis-aligned rules, attackers can be more realistically modelled with asymmetric changes and different constraints for different axes. Also, attackers can be modelled with a `budget' that they can spend on changing data points which allows the user to evaluate robustness against attackers of different strengths. TREANT directly optimizes a loss function instead of using a splitting criterion like in regular or robust decision trees \cite{chen_robust_2019}. Although this allows TREANT to train robust models against a variety of attackers, their algorithm deploys a solver to optimize the loss function and pre-computes all possible attacks in exponential time.

\section{Specifying Threat Models} \label{sec:threat-model}
In our work we assume existence of an attacker that knows the decision tree and perturbs samples according to a user-specified threat model. Therefore to support a wide range of attack types we take inspiration from TREANT \cite{calzavara_treant_2019} and let the user define the perturbation limits for each individual feature. The specification is as follows:
\begin{itemize}
    \item ``''or None: This feature cannot be perturbed.
    \item $>$ or $<$: This feature can be increased / decreased.
    \item $<>$: This feature can be perturbed to any value.
    \item $\epsilon$: The feature can be perturbed by a distance of $\epsilon$.
    \item ($\epsilon_l$, $\epsilon_r$): The feature can be perturbed $\epsilon_l$ left or $\epsilon_r$ right.
\end{itemize}
It is worth noting that all these cases can be translated to the tuple notation, e.g. we can encode $>$ as $(0, \infty)$ or a number $\epsilon$ as $(\epsilon, \epsilon)$. For conciseness we only use the tuple notation in the algorithms in section \ref{sec:groot}. When we set the threat model to $\epsilon$ for each feature, it behaves identically to an $L_\infty$ norm. In that case \textsc{groot} performs similarly to the Gini impurity variant of algorithm 1 by Chen et al. \cite{chen_robust_2019}, with only changes to the number of candidate splits and what samples propagate left and right. Our implementation of \textsc{GROOT} and adversarial attacks can be found on GitHub\footnote{\url{https://github.com/tudelft-cda-lab/GROOT}}.

\begin{figure}[t!]
    \centering
    \begin{subfigure}[b]{0.4\linewidth}
        \centering
        \includegraphics[width=\linewidth]{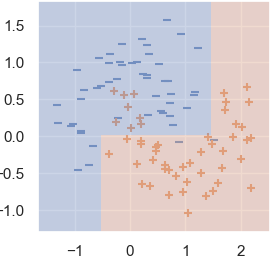}
        \caption{No attacker}
        \label{fig:threat-model-vizualization-none}
    \end{subfigure}
    \begin{subfigure}[b]{0.4\linewidth}
        \centering
        \includegraphics[width=\linewidth]{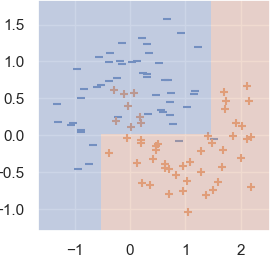}
        \caption{$(>, <)$}
        \label{fig:threat-model-vizualization-direction}
    \end{subfigure}
    \newline
    \begin{subfigure}[b]{0.4\linewidth}
        \centering
        \includegraphics[width=\linewidth]{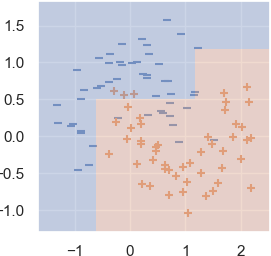}
        \caption{$(0.5, 0.5)$}
        \label{fig:threat-model-vizualization-linf}
    \end{subfigure}
    \begin{subfigure}[b]{0.4\linewidth}
        \centering
        \includegraphics[width=\linewidth]{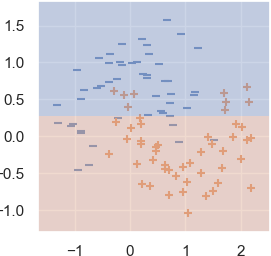}
        \caption{$(<>, (0.7, 0.3))$}
        \label{fig:threat-model-vizualization-complex}
    \end{subfigure}
    \newline
    \caption{Decision regions of \textsc{groot} trees attacked by different threat models (indicated below each image). Attacker only move the orange samples. The threat model greatly influences the learned trees. For instance, robust decision trees against $L_\infty$ perturbations (\ref{fig:threat-model-vizualization-linf}) are very different from trees robust against more complex attackers (\ref{fig:threat-model-vizualization-direction}, \ref{fig:threat-model-vizualization-complex}).}
    \label{fig:threat-models-visualization}
\end{figure}

In this work we assume that we train models to distinguish between two classes. In contrast to previous works we allow one to select whether the attacker can move all samples or only samples of one class. We do this since in security settings it would intuitively be impractical for a user's benign data to be predicted as malicious. Take for example a spam email detector, a benign user would not want their emails to be predicted as spam while a spammer would want their emails to be mistaken as benign. We give examples of threat models and highlight their importance in Figure \ref{fig:threat-models-visualization}.

\section{Adversarial Gini Impurity} \label{sec:adversarial-gini-impurity}
\subsection{Adversarial Gini Impurity for Two Moving Classes}
We typically fit decision trees with a splitting criterion such as the Gini impurity. To determine the quality of a split we then take the weighted average of the scores on both sides. We can define the Gini impurity for two classes as:

\small
\begin{equation}
    G(N_0, N_1) = 1 - \left(\tfrac{N_0}{N_0 + N_1}\right)^2 - \left(\tfrac{N_1}{N_0 + N_1}\right)^2
\end{equation}
\normalsize

Where $N_0$ and $N_1$ are the number of samples of label $0$ and $1$ respectively. Then we combine this into a score function by taking the weighted average with respect to number of samples on each side of the split (other works use the Gini gain which behaves identically):

\small
\begin{multline} \label{eq:score-function-gini}
    S(L_0, L_1, R_0, R_1) =\\
    \frac{(L_0 + L_1) \cdot G(L_0, L_1) + (R_0 + R_1) \cdot G(R_0, R_1)}{L_0 + L_1 + R_0 + R_1}
\end{multline}
\normalsize

Where $L_0$ and $L_1$ are the number of samples on the left side of the split of label $0$ (benign) and $1$ (malicious) respectively. Similarly $R_0$ and $R_1$ represent samples on the right. Normally one searches for a split that minimizes this score function. Instead, we minimize the score function after worst-case influence by an attacker.

\begin{figure}[b!]
    \centering
    \includegraphics[width=0.9\linewidth]{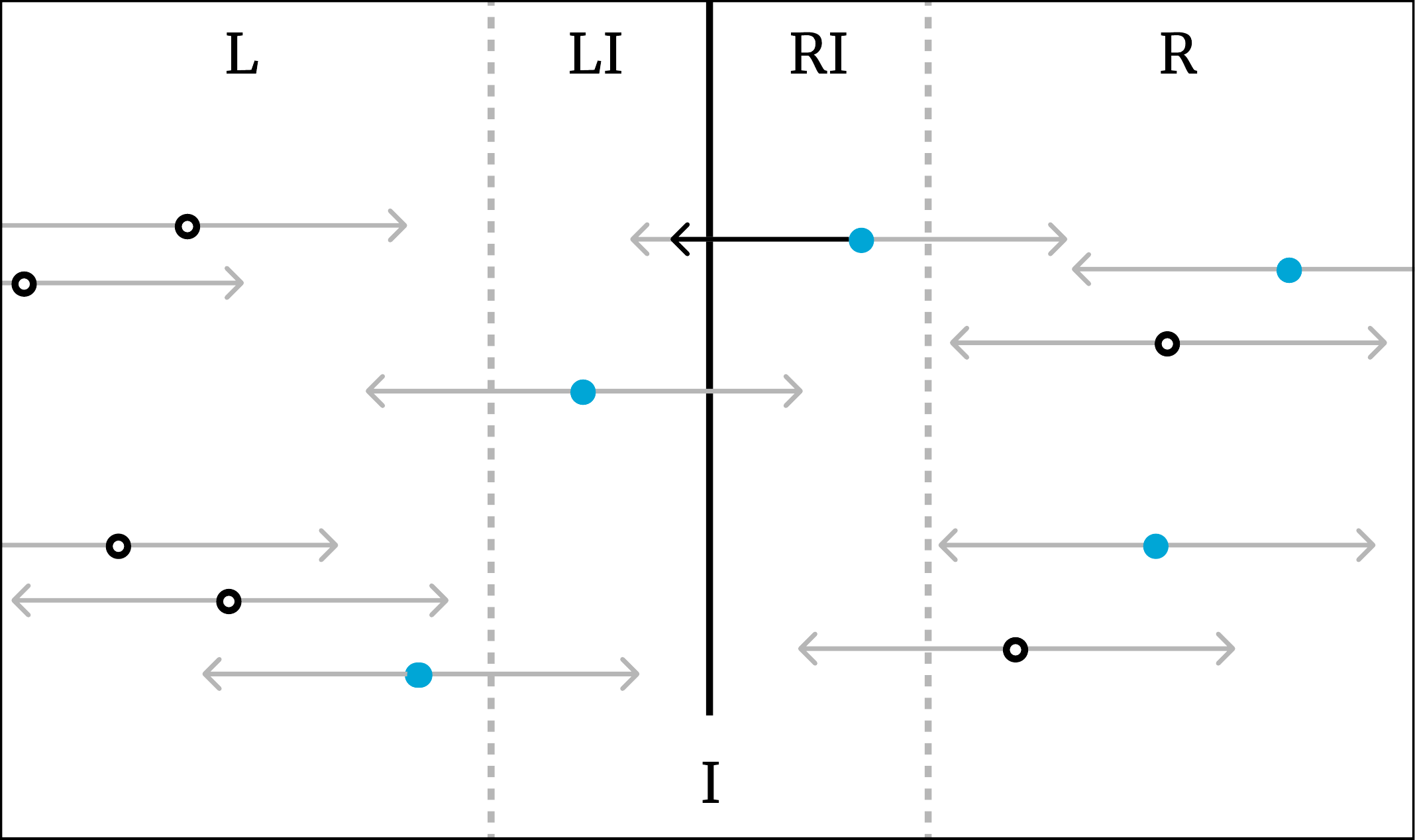}
    \caption{Example of the adversarial Gini impurity where samples can move in the range of the arrows. We want to move a number of samples from $I$ over the threshold (line, center) to maximize the weighted average of Gini impurities. In this example we can move the single blue (filled) sample from $RI$ into $LI$ to maximize it.}
    \label{fig:adversarial-gini}
\end{figure}

Where one normally minimizes the Gini impurity, we assume an attacker that aims to maximize $S(\cdot)$ by moving samples from $I$ to different sides of the split. We visualize this maximization problem in Figure \ref{fig:adversarial-gini}. Here, $I_1$ is the number of points with label $1$ (malicious) that are close enough to the split that the adversary can move them to either side. Mathematically we are looking for the integer $x \in [0, I_1]$ such that $x$ points of $I_1$ move to the left side of the split and $I_1 - x$ to the right. Similarly we have a $I_0$ and $y$ for the benign samples. The score function under attacker influence is:

\small
\begin{multline} \label{eq:robust-score-function}
    S_{\text{robust}}(L_0, L_1, R_0, R_1, I_0, I_1) = \max_{x \in [0, I_1], y \in [0, I_0]}\\
    S(L_0 + y, L_1 + x, R_0 + I_0 - y, R_1 + I_1 - x)
\end{multline}
\normalsize

\noindent We can then write the $x'$ and $y'$ that maximize it as:

\small
\begin{multline} \label{eq:maximize_score_function}
    x', y' = \argmax_{x \in [0, I_1]} \left(
        \frac{(L_0 + y) (L_1 + x)}{L_0 + L_1 + x + y}\right.\\
        \left.+
        \frac{(R_0 + I_0 - y) (R_1 + I_1 - x)}{R_0 + R_1 + I_0 + I_1 - x - y}
    \right)
\end{multline}
\normalsize

The algorithm by Chen et al. \cite{chen_robust_2019} optimizes a similar function by iterating through the $I$ samples to perform gradient ascent. However, since the function is concave with respect to $x$ and $y$, we can do this faster by maximizing the function analytically and rounding to a near integer solution. The maxima form the following line (proofs in the appendix):

\small
\begin{equation} \label{eq:solution-line}
    y' = \frac{L_1 (R_0 + I_0) - L_0 (R_1 + I_1)}{L_1 + R_1 + I_1}
    + \frac{(L_0 + R_0 + I_0) x'}{L_1 + R_1 + I_1}
\end{equation}
\normalsize

Using gradient ascent and given enough movable samples one would end up on the optimal line but not necessarily on the closest point on the line to the starting values for $x$ and $y$. We argue that the closest point is intuitively a better solution than a random point on the line, because it represents the least number of samples to move. Therefore in \textsc{groot} we find the closest point on the solution line to the starting values $x, y$ then round to the nearest integers.

Computing the point and rounding it all takes $\mathcal{O}(1)$ time. Therefore we have an efficient method (constant time) to compute $S_{\text{robust}}$

\subsection{Adversarial Gini Impurity for Malicious Samples}
In \textsc{groot} we intent to allow users to choose whether all samples can move or only malicious samples. When only malicious samples move, computing the adversarial Gini impurity becomes simpler. In that case we can rewrite Equation \ref{eq:solution-line} in terms of $y$, then set $y$ and $I_0$ to $0$ since benign samples do not move:

\small
\begin{equation}
    x': \frac{L_0 R_1 + L_0 I_1 - L_1 R_0}{L_0 + R_0}
\end{equation}
\normalsize

This solution can again be computed in constant time then rounded to $\lfloor x' \rfloor$ and $\lceil x' \rceil$ to find the integer maximum of $x$.

\section{\textsc{groot}} \label{sec:groot}
We introduce \textsc{groot} (Growing RObust Trees), an algorithm that trains decision trees that are robust against adversarial examples generated from a user-specified threat model. The algorithm stays close to regular decision tree learning algorithms but searches through more candidate splits with a robust score function and propagates samples according to an attacker. Like regular decision tree learning algorithms, \textsc{groot} runs in $\mathcal{O}(n \log n)$ time in terms of $n$ samples.

\subsection{Scoring Candidate Splits}
Similar to regular decision tree learning algorithms we can search over all possible splits and compute a score function to find the best split. In Algorithm \ref{alg:robust-split} we iterate over each sample in sorted order to identify candidate splits. We evaluate each candidate split with the adversarial Gini impurity from the previous subsection to find the split that is most accurate against an adversary. To support both numerical and categorical features we split the search up in the two cases and describe each of these below. The time complexity in terms of $n$ samples for both cases is bounded by $\mathcal{O}(n \log{n})$ per feature.

\paragraph{Numerical case}
In regular decision tree learning algorithms we score candidate splits at each position in which a sample moves from the right to the left side. However, when an adversary can perturb samples there are more possible splits that affect the sample counts on each side. Therefore we consider also candidate splits where a movable sample becomes in or out of range of $I$. Take for example a sample at position $3.3$ that can be perturbed in a radius of $0.2$, we score a split at $3.1$, $3.3$ and $3.5$. At the start of Algorithm \ref{alg:robust-split} we sort all candidate splits and store what will happen to the counts $L$ / $R$ / $I$ at those splits, e.g. a sample moves from $R$ to $RI$. We can evaluate splits in $\mathcal{O}(1)$ time as explained in Section \ref{sec:adversarial-gini-impurity} and we consider at maximum $3n$ splits, where $n$ is number of samples. Therefore the time complexity of evaluating splits is $\mathcal{O}(n)$ per feature and this means the fitting run time is dominated by the sorts of complexity $\mathcal{O}(n\log{n})$.

\paragraph{Categorical case}
To support categorical variables, we must find the best partition of categories with respect to the scoring function. Although a linear time algorithm (in number of categories) exists for fitting regular decision trees \cite{coppersmith_nominal_1999}, we have not found such an algorithm for robust decision trees. Since the number of categories in a feature is usually bounded by a small constant, we perform an exponential time search over all possible partitions. We can count the number of samples of each category in linear time, so the complexity for this procedure is $\mathcal{O}(n 2^c)$ where $n$ is the number of samples and $c$ the number of categories. We do not explicitly give an algorithm as it follows the same idea as Algorithm \ref{alg:robust-split}.

\begin{algorithm}[t!]
    \caption{Find Best Robust Split on Numerical variable}\label{alg:robust-split}
    \begin{algorithmic}[1]
        \Function{BestRobustSplit}{$X$}
            \State $S \gets X \cup \{o - \epsilon_l | o \in X\} \cup \{o + \epsilon_r | o \in X\}$
            \For{$s \in S$}
                \State $R \gets \{o | o \in X \land o > s + \epsilon_r\}$
                \State $RI \gets \{o | o \in X \land s < o \leq s + \epsilon_r\}$
                \State $LI \gets \{o | o \in X \land s - \epsilon_l < o \leq s\}$
                \State $L \gets \{o | o \in X \land o \leq s - \epsilon_l\}$
                \State $l_0 \gets |L_0| + (1 - \rho) |LI_0|$
                \State $l_1 \gets |L_1| + (1 - \rho) |LI_1|$
                \State $r_0 \gets |R_0| + (1 - \rho) |RI_0|$
                \State $r_1 \gets |R_1| + (1 - \rho) |RI_1|$
                \State $i_0 \gets \rho |RI_0| + \rho |LI_0|$,\: $i_1 \gets \rho |RI_1| + \rho |LI_1|$
                \State $(x_s, y_s) \gets$ close point on Eq. \ref{eq:solution-line} to  $(|LI_0|, |LI_1|)$
                \State $x_s \gets \lfloor x_s \rceil, \quad y_s \gets \lfloor y_s \rceil$
                \State $g_{s} \gets S(l_0 + y_s,\ l_1 + x_s,\ r_0 + i_0 - y_s,\ r_1 + i_1 - x_s)$
            \EndFor
            \State $s' \gets \argmin_{s}\ g_{s}$
            \State \Return $(s', g_{s'}, x_{s'})$
        \EndFunction
    \end{algorithmic}
\end{algorithm}

\subsection{Propagating Samples}
When fitting regular decision trees one can simply move all samples lower than a threshold left and higher to the right. To account for samples that the adversary moves we make a modification to this propagation as we define in Algorithm \ref{alg:robust-fit}. We do not only keeping track of left ($L$) and right ($R$) samples, but also store `intersection' sets ($RI$, $LI$) that contains samples that can move to both sides.

In section \ref{sec:adversarial-gini-impurity} we showed that $x'$ and $y'$ are the optimal values for the adversarial Gini impurity. Given $x'$ we move samples from $I_1$ over the split to place $x'$ samples on the left and $I_1 - x'$ on the right. If there were fewer than $x'$ samples on the left we move samples from the right, if there were more samples on the left we move them to the right. The actual samples that move are randomly selected from the intersection. If the user selects that the attacker can move all classes we repeat this operation for $I_0$ and $y'$ to account for moving benign samples.

\begin{algorithm}[t!]
    \caption{Fit Robust Tree on Numerical Data}\label{alg:robust-fit}
    \begin{algorithmic}[1]
        \Function{FitRobustTree}{X}
            \If{stopping criterion (e.g. maximum depth)}
                \State \Return $\text{Leaf}(|X_0|, |X_1|)$
            \Else
                \For{$f \gets 1 ... F$}
                    \State $s_f, g_f, x_f \gets \Call{BestRobustSplit}{X^f}$
                \EndFor
                \State $f' \gets \argmin_{f}\ g_f$
                \State determine $R$, $RI$, $LI$, $L$ for split $f'$
                \State move $(1 - \rho)|LI|$ samples from $LI$ to $L$
                \State move $(1 - \rho)|RI|$ samples from $RI$ to $R$
                \If{$|LI| < x_{f'}$}
                    \State move $|LI| - x_{f'}$ samples from $RI$ to $LI$
                \Else
                    \State move $x_{f'} - |LI|$ samples from $LI$ to $RI$
                \EndIf
                \State $\text{node}_l \gets \Call{FitRobustTree}{L \cup LI}$
                \State $\text{node}_r \gets \Call{FitRobustTree}{R \cup RI}$
                \State \Return $\text{DecisionNode}(s_{f'}, \text{node}_l, \text{node}_r)$
            \EndIf
        \EndFunction
    \end{algorithmic}
\end{algorithm}

\subsection{Trading Off Regular and Adversarial Performance}

When maximizing performance in an adversarial setting one sacrifices performance in the regular unperturbed setting. To trade off between performance in these settings, we introduce a parameter $0 \leq \rho \leq 1$ that determines the fraction of malicious points that can move. Intuitively this encodes what percentage of attackers attempt an evasion attack. For example, if we deploy a robust model for malware detection, we can expect that only a fraction of malware creators will change their code to evade the model.

We make two changes to the algorithms to account for $\rho$. First, we scale the variable $i$ in Algorithm \ref{alg:robust-split} to $\rho\, i$. Second, during sample propagation (Algorithm \ref{alg:robust-fit}) we reduce the size of the set $I$ to $\rho |I|$ by randomly returning $(1-\rho) |I|$ samples to their unperturbed side of the split ($L$ or $R$).

\section{Results} \label{sec:results}
We give results of regular (natural) decision trees, TREANT \cite{calzavara_treant_2019}, Chen et al. \cite{chen_robust_2019} and \textsc{groot} on fourteen datasets in which we compare predictive performance and run time.
For natural decision trees we use scikit-learn's \cite{scikit-learn} implementation as it is widely used for research in the field.
To compare against robust decision tree algorithms we run TREANT and the heuristic from Chen et al. that computes the adversarial Gini impurity using four representative cases. All datasets can be retrieved from openML except breast-cancer, their specific versions are listed in the tables. Breast-cancer was imported from Scikit-learn.

\begin{table*}[t!]
    \centering
    \setlength{\tabcolsep}{7.6pt}
    \caption{
    Test accuracy scores after adversarial attacks, best adversarial scores are bold. Mean and standard deviations were computed over 5 fold cross validation. An adversarial attack succeeded when the decision tree had a leaf of different prediction label within $L_\infty$ distance $0.1$.} \label{tab:results-two-classes}
    \begin{tabular}{l|ll|cccc}
    \toprule
    Dataset & Samples & Features & Chen et al. & GROOT & Natural & TREANT \\
    \midrule
    banknote-authentication (1) & 1372 & 4 & .929 $\;\pm\;$ .020 & \textbf{.943} $\;\pm\;$ .017 & .930 $\;\pm\;$ .034 & .938 $\;\pm\;$ .027 \\
    blood-transfusion (1) & 748 & 4 & \textbf{.769} $\;\pm\;$ .033 & \textbf{.769} $\;\pm\;$ .033 & \textbf{.769} $\;\pm\;$ .033 & .762 $\;\pm\;$ .012 \\
    breast-cancer & 569 & 30 & \textbf{.926} $\;\pm\;$ .013 & \textbf{.926} $\;\pm\;$ .013 & .341 $\;\pm\;$ .103 & \textbf{.926} $\;\pm\;$ .014 \\
    climate-model-simulation (4) & 540 & 18 & .844 $\;\pm\;$ .044 & .902 $\;\pm\;$ .012 & .748 $\;\pm\;$ .041 & \textbf{.915} $\;\pm\;$ .004 \\
    cylinder-bands (2) & 277 & 37 & \textbf{.664} $\;\pm\;$ .076 & \textbf{.664} $\;\pm\;$ .076 & .603 $\;\pm\;$ .037 & .621 $\;\pm\;$ .085 \\
    diabetes (1) & 768 & 8 & .729 $\;\pm\;$ .025 & .727 $\;\pm\;$ .016 & .720 $\;\pm\;$ .020 & \textbf{.732} $\;\pm\;$ .029 \\
    haberman (1) & 306 & 3 & \textbf{.722} $\;\pm\;$ .046 & \textbf{.722} $\;\pm\;$ .046 & \textbf{.722} $\;\pm\;$ .046 & .706 $\;\pm\;$ .038 \\
    ionosphere (1) & 351 & 34 & \textbf{.880} $\;\pm\;$ .016 & .872 $\;\pm\;$ .010 & .712 $\;\pm\;$ .048 & .869 $\;\pm\;$ .031 \\
    parkinsons (1) & 195 & 22 & .800 $\;\pm\;$ .066 & \textbf{.826} $\;\pm\;$ .046 & .174 $\;\pm\;$ .084 & .821 $\;\pm\;$ .065 \\
    planning-relax (1) & 182 & 12 & .655 $\;\pm\;$ .067 & .676 $\;\pm\;$ .043 & .468 $\;\pm\;$ .091 & \textbf{.709} $\;\pm\;$ .013 \\
    sonar (1) & 208 & 60 & .336 $\;\pm\;$ .096 & .432 $\;\pm\;$ .083 & .049 $\;\pm\;$ .096 & \textbf{.500} $\;\pm\;$ .073 \\
    spambase (1) & 4601 & 57 & .843 $\;\pm\;$ .014 & \textbf{.874} $\;\pm\;$ .019 & .340 $\;\pm\;$ .061 & .837 $\;\pm\;$ .014 \\
    SPECTF & 267 & 44 & \textbf{.753} $\;\pm\;$ .059 & \textbf{.753} $\;\pm\;$ .059 & .749 $\;\pm\;$ .037 & .719 $\;\pm\;$ .058 \\
    wine (1) & 6497 & 11 & .635 $\;\pm\;$ .043 & \textbf{.681} $\;\pm\;$ .012 & .500 $\;\pm\;$ .057 & .670 $\;\pm\;$ .016 \\
    \bottomrule
    \end{tabular}
\end{table*}

\subsection{Predictive performance}
To determine the quality of the trees produced by each algorithm we measure predictive performance in a regular and adversarial setting. When measuring in the adversarial setting, an attacker first optimally moves all test samples within the perturbation limits to maximize the number of misclassifications. To maximize the number of misclassifications within these limits we loop over all tree leaves and check if the leaf is in reach of the sample. If any reachable leaf predicts a different class than the actual label we count a misclassification. The adversarial accuracy is the accuracy after this procedure or equivalently the success rate of an attacker given the perturbation limits. In the experiments we test against an attacker that can move samples freely within $L_\infty$ distance $\epsilon = 0.1$.

We encoded the above threat models in TREANT's attack rules using precondition $[-\infty, \infty]$ and postcondition $\epsilon$ or $-\epsilon$. Each rule has cost $1$ and the attacker has a budget equal to the depth of the trees.
To exactly match the threat model specifications, we modify TREANT to only use attack rules once per feature. Preliminary testing without this modification gave poor results. Given these modifications the attack rules exactly encode the $L_\infty$ radius attack model.

A popular method for measuring robustness (e.g. in \cite{chen_robust_2019}) is to compute the average perturbation distance required to cause a misclassification. We choose against this metric as it assumes features can perturb arbitrarily and with equal cost.

\begin{figure}[t!]
    \centering
    \includegraphics[width=0.99\linewidth]{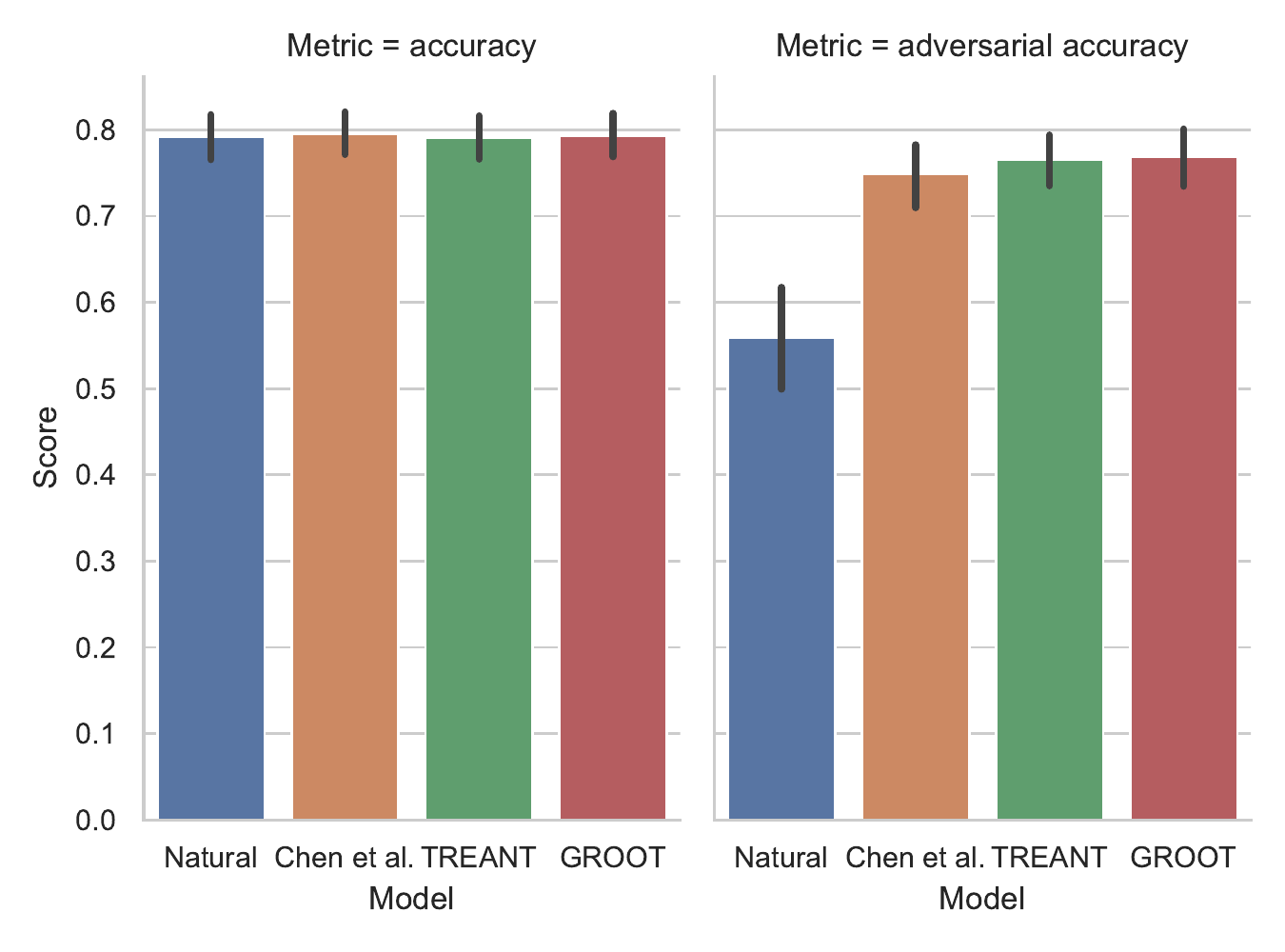}
    \caption{Average accuracy scores over 14 datasets before and after adversarial attacks. All models fit accurate trees while TREANT and \textsc{GROOT} perform best in the adversarial setting.}
    \label{fig:average-score}
\end{figure}

We train a decision tree on each dataset with 5 fold stratified cross validation. All algorithms were trained up to a depth of $4$ to maintain interpretability. We implemented GROOT and the heuristic by Chen et al. in python but used TREANT's existing implementation with minor modifications. Scikit-learn uses the Gini impurity and TREANT optimizes the sum of squared errors. We used $\rho = 1$ for \textsc{GROOT} to maximize adversarial performance. We used the default hyperparameter settings from Scikit-learn for all models which means leaves needed 2 samples to be created and trees could split multiple times on the same feature. We present results on adversarial accuracy in Table \ref{tab:results-two-classes} and summarize the results on accuracy and adversarial accuracy in Figure \ref{fig:average-score}.

While all four models considered scored well on accuracy, the scores in the adversarial setting differ. TREANT and \textsc{groot} perform similarly on adversarial accuracy and both significantly improve on Scikit-learn's natural decision tree scores by approximately $21\%$. The heuristic by Chen et al. score approximately $2\%$ worse than TREANT and \textsc{GROOT}.
The cases where TREANT outperforms \textsc{groot} and Chen could be explained by TREANT's attack invariance that prevents deeper trees from becoming less robust. The heuristic by Chen et al. finds an improvement over \textsc{GROOT} for one dataset with a $0.8\%$ improvement in adversarial accuracy. We expect this result to be insignificant as it is within the range of one standard deviation.

On the blood-transfusion and haberman datasets the natural decision trees perform equally well on adversarial accuracy as the robust tree fitting algorithms. These datasets can be well predicted using features that have samples that lie far apart i.e. more than $0.1$ distance.

\subsection{Run time}
\begin{figure}[t!]
    \centering
    \includegraphics[width=0.99\linewidth]{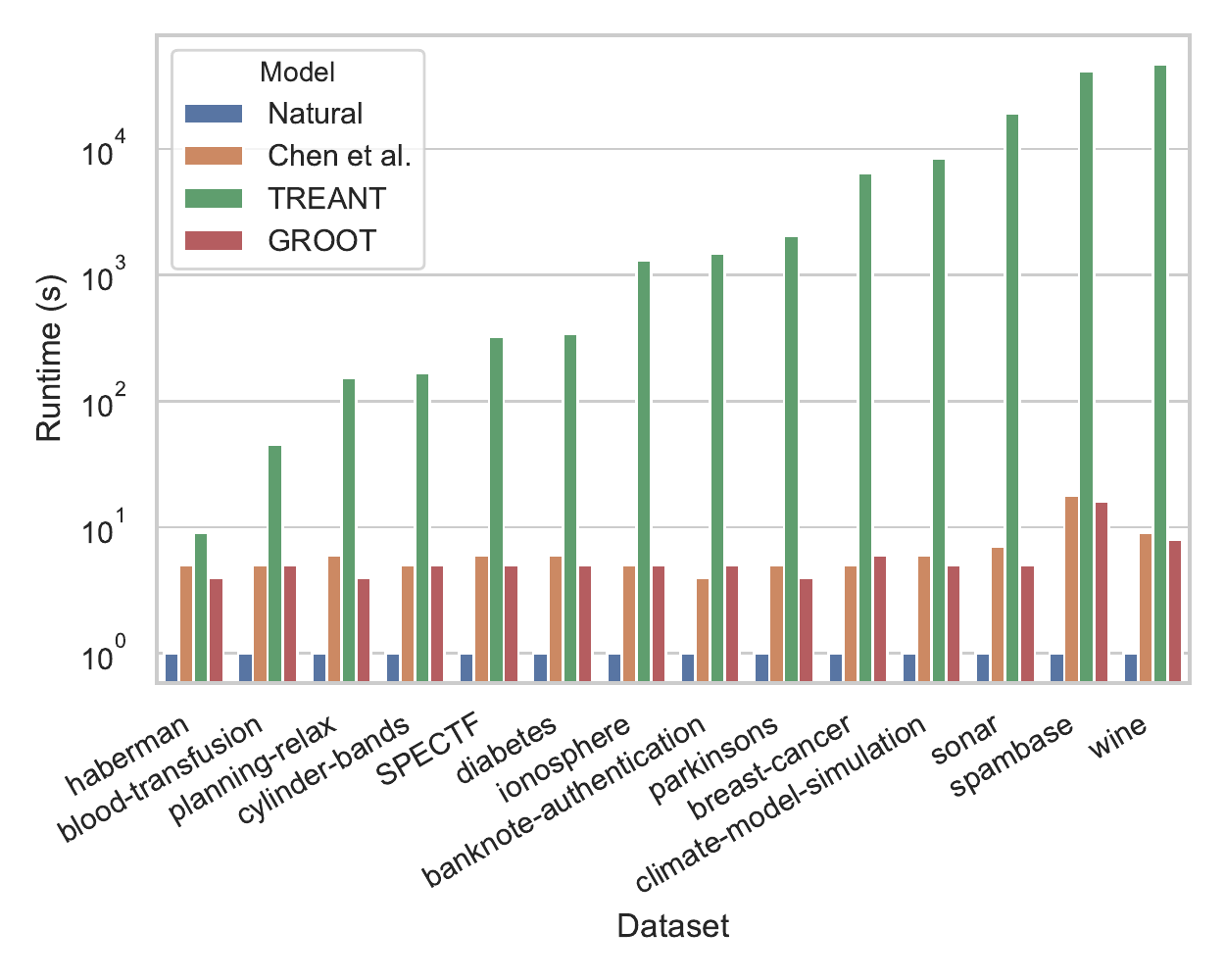}
    \caption{Runtimes in seconds on a logaritmic scale for fitting decision trees of depth 4. Natural trees use Scikit-learn's fast implementation, Chen et al. and GROOT ran up to a thousand times faster than TREANT.}
    \label{fig:runtime}
\end{figure}

To compare the efficiency of the algorithms, we also recorded the run times in Figure \ref{fig:runtime}. All experiments ran on a linux machine with 16 Intel Xeon CPU cores and 72GB of RAM total. Each algorithm instance ran on a single core.

Although the run time complexities for scikit-learn and our work are similar, the efficient implementation of scikit-learn runs fast even for large sample sizes. It is worth noting that \textsc{groot} cannot be faster than scikit-learn as \textsc{groot} searches over more candidate splits. Comparing TREANT and \textsc{groot}, we see that our algorithm runs on average two orders of magnitude faster. TREANT exhaustively searches for attacks using an exponential search and uses a sequential quadratic programming solver to optimize the loss function which likely contributes to the higher run time. The heuristic by Chen et al. has a similar runtime as GROOT as it is implemented in the same python code and shares the same time complexity.

\section{Discussion and conclusions} \label{sec:discussion-conclusion}
We present \textsc{groot}, a greedy algorithm for learning robust decision trees. It uses an analytical solution for computing the adversarial Gini impurity and runs two orders of magnitude faster than existing approaches. Our results show that \textsc{groot} runs in a matter of seconds and is competitive with the state-of-the-art TREANT method.

An interesting direction for future work is to investigate the performance of \textsc{groot} in a black box setting. Specifically, we are interested in the susceptibility to boundary and transfer attacks. As \textsc{groot} was designed for classification between benign and malicious data points it currently does not support regression tasks. However, decision tree learners can be modified to work on continuous target values. We aim to investigate such modifications in future work, as well as the performance in random forests and gradient boosting ensembles.

For the sake of comparison, we experimented using the same public datasets as similar works on robustness, but these datasets were originally not intended for research into adversarial attacks. With \textsc{groot}'s flexible implementation, it becomes possible to model different kinds of attack on different kinds of datasets, bringing adversarial machine learning studies closer to the real world. In the near future, we will apply this framework to datasets from problems where adversarial modifications are an important concern such as fraud and intrusion detection.

\noindent We conclude that:
\begin{itemize}
    \item By solving the adversarial Gini impurity analytically we can now fit robust trees with the same time complexity as regular trees: $\mathcal{O}(n \log n)$, for $n$ samples. 
    \item This algorithm runs two orders of magnitude faster than the state-of-the-art work and as efficiently as a fast heuristic.
    \item \textsc{groot} consistently achieves scores competitive with the state-of-the-art work in terms of robustness.
\end{itemize}

\bibliographystyle{named}
\bibliography{references}

\cleardoublepage

\section*{Appendix: Adversarial Gini Impurity Proofs and Solutions}
In Section 4 we introduced the adversarial Gini impurity, a score function that promotes robust splits. This function consists of a  maximization term that is dependant on two variables: $x$ and $y$. Although we identified the analytical solutions for $x$ and $y$ we did not prove they were optimal or show how to find the solutions. To show the solutions are optimal we first prove that the maximization term of the adversarial Gini impurity is concave with respect to $x$ and $y$. Given that this term is concave we then find the maxima by identifying the critical points that fall inside our region of interest.

\subsection*{Proof of Concavity}
Let us prove that the maximization term of the adversarial Gini impurity is concave so that critical points are maxima. The maximization term is:

\begin{equation*}
    f(x, y) =
    \frac{(L_0 + y) (L_1 + x)}{L_0 + L_1 + x + y}
        +
    \frac{(R_0 + I_0 - y) (R_1 + I_1 - x)}{R_0 + R_1 + I_0 + I_1 - x - y}
\end{equation*}

\begin{figure*}
    \centering
    \includegraphics[width=.95\linewidth]{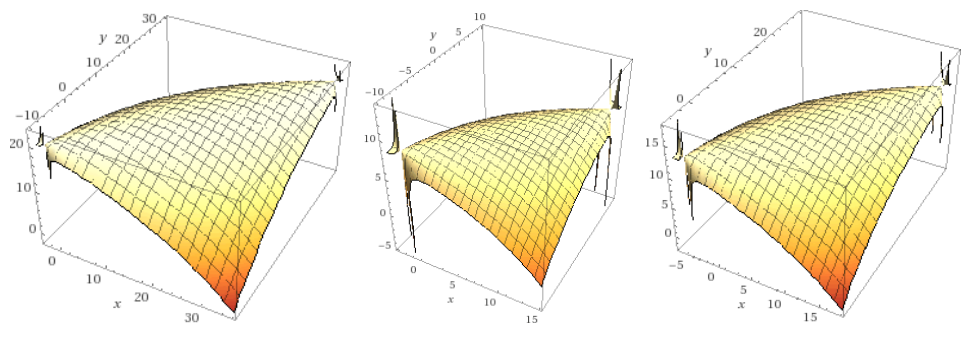}
    \caption{Function that we attempt to maximize for different settings of $L_0$, $L_1$, $R_0$, $R_1$, $I_0$ and $I_1$, plotted against $x$ and $y$. The maxima appear to be lines through the diagonals of the plots.}
    \label{fig:concavity-intuition}
\end{figure*}

We can get an intuition for $f$'s concavity by plotting it for some random choices of $L$, $R$ and $I$, see Figure \ref{fig:concavity-intuition}.

A function is concave if its hessian is negative semi definite. To show that the hessian is negative semi definite we can use Sylvester's criterion. This criterion states that if the even order principal minors of a hermitian matrix are non-negative and the odd order principal minors non-positive then the matrix is negative semi definite. Since the hessian of a real function with two variables ($x, y$) is symmetric and real, it is hermitian. Therefore we can apply Sylvester's criterion here. Let us first compute the hessian:

\[
    H = \begin{bmatrix}
    \frac{\partial}{\partial x}\left( \frac{\partial}{\partial x} f(x, y) \right) &
    \frac{\partial}{\partial x}\left( \frac{\partial}{\partial y} f(x, y) \right) \\
    \frac{\partial}{\partial y}\left( \frac{\partial}{\partial x} f(x, y) \right) &
    \frac{\partial}{\partial y}\left( \frac{\partial}{\partial y} f(x, y) \right)
    \end{bmatrix}
\]

Let us work out each term:

\scriptsize
\begin{gather*}
    \frac{\partial}{\partial x}\left( \frac{\partial}{\partial x} f(x, y) \right) = \\
    - 2 \frac{L_0 + y}{(L_0 + L_1 + x +y)^2} + 2 \frac{(L_0 + y) (L_1 + x)}{(L_0 + L_1 + x + y)^3}\\
    - 2 \frac{R_0 + I_0 - y}{(R_0 + R_1 + I_0 + I_1 - x - y)^2} + 2 \frac{(R_0 + I_0 - y) (R_1 + I_1 - x)}{(R_0 + R_1 + I_0 + I_1 - x - y)^3} = \\
    -2 \frac{(L_0 + y)^2}{(L_0 + L_1 + x + y)^3} - 2 \frac{(R_0 + I_0 - y)^2}{(R_0 + R_1 + I_0 + I_1 - x - y)^3}
\end{gather*}

\begin{gather*}
    \frac{\partial}{\partial y}\left( \frac{\partial}{\partial y} f(x, y) \right) = \\
    - 2 \frac{L_1 + x}{(L_0 + L_1 + x +y)^2} + 2 \frac{(L_0 + y) (L_1 + x)}{(L_0 + L_1 + x + y)^3}\\
    - 2 \frac{R_1 + I_1 - x}{(R_0 + R_1 + I_0 + I_1 - x - y)^2} + 2 \frac{(R_0 + I_0 - y) (R_1 + I_1 - x)}{(R_0 + R_1 + I_0 + I_1 - x - y)^3} = \\
    -2 \frac{(L_1 + x)^2}{(L_0 + L_1 + x + y)^3} - 2 \frac{(R_1 + I_1 - x)^2}{(R_0 + R_1 + I_0 + I_1 - x - y)^3}
\end{gather*}
\normalsize

Since the order of differentiation does not matter, we have that: $\frac{\partial}{\partial x}\left( \frac{\partial}{\partial y} f(x, y) \right) = \frac{\partial}{\partial y}\left( \frac{\partial}{\partial x} f(x, y) \right)$. Therefore we only have to compute one of the sides of this equation.

\scriptsize
\begin{gather*}
    \frac{\partial}{\partial x}\left( \frac{\partial}{\partial y} f(x, y) \right) = \frac{\partial}{\partial y}\left( \frac{\partial}{\partial x} f(x, y) \right) = \\
    \frac{1}{L_0 + L_1 + x + y} - \frac{L_0 + y}{(L_0 + L_1 + x + y)^2} - \frac{L_1 + x}{(L_0 + L_1 + x + y)^2}\\
    + 2 \frac{(L_0 + y) (L_1 + x)}{(L_0 + L_1 + x +y)^3} + \frac{1}{R_0 + R_1 + I_0 + I_1 - x - y}\\
    - \frac{R_0 + I_0 - y}{(R_0 + R_1 + I_0 + I_1 - x - y)^2} - \frac{R_1 + I_1 - x}{(R_0 + R_1 + I_0 + I_1 - x - y)^2} \\
    + 2 \frac{(R_0 + I_0 - y) (R_1 + I_1 - x)}{(R_0 + R_1 + I_0 + I_1 - x - y)^3} = \\
    2 \frac{(L_0 + y) (L_1 + x)}{(L_0 + L_1 + x +y)^3} + 2 \frac{(R_0 + I_0 - y) (R_1 + I_1 - x)}{(R_0 + R_1 + I_0 + I_1 - x - y)^3}
\end{gather*}
\normalsize

Now let us refer back to Sylvester's criterion, we want to show that the first principal minor is non-positive and the second non-negative. The first principal minor is simply the term in the top-left corner of $H$:

\begin{equation*}
    -2 \frac{(L_0 + y)^2}{(L_0 + L_1 + x + y)^3} - 2 \frac{(R_0 + I_0 - y)^2}{(R_0 + R_1 + I_0 + I_1 - x - y)^3}
\end{equation*}

Since the terms $L_0$, $L_1$, $R_0$, $R_1$ and $I_1 - x$, $I_0 - y$ are all non-negative, both fractions are non-negative. The $-2$ in front of them then make the both non-positive. Therefore for our region of interest the first principal minor is non-positive.

Now we continue with the second principal minor, this is the determinant of $H$. For simplicity let us label the four elements of $H$:

$$H = \begin{bmatrix} a & b \\ c & d \end{bmatrix}$$

The second principal minor is then $\det{H} = ad - bc$. We will show that this determinant is positive:

\small
\begin{gather*}
    \left(\frac{(L_0 + y)^2}{(L_0 + L_1 + x + y)^3} + \frac{(R_0 + I_0 - y)^2}{(R_0 + R_1 + I_0 + I_1 - x - y)^3} \right) \\
    \left( \frac{(L_1 + x)^2}{(L_0 + L_1 + x + y)^3} + \frac{(R_1 + I_1 - x)^2}{(R_0 + R_1 + I_0 + I_1 - x - y)^3} \right) - \\
    \left( \frac{(L_0 + y) (L_1 + x)}{(L_0 + L_1 + x +y)^3} + \frac{(R_0 + I_0 - y) (R_1 + I_1 - x)}{(R_0 + R_1 + I_0 + I_1 - x - y)^3} \right) \\
    \left( \frac{(L_0 + y) (L_1 + x)}{(L_0 + L_1 + x +y)^3} + \frac{(R_0 + I_0 - y) (R_1 + I_1 - x)}{(R_0 + R_1 + I_0 + I_1 - x - y)^3} \right) = \\
    \left( \frac{(L_0 + y)^2 (R_1 + I_1 - x)^2 + (L_1 + x)^2(R_0 + I_0 - y)^2}{(L_0 + L_1 + x + y)^3 (R_0 + R_1 + I_0 + I_1 - x - y)^3} \right) - \\
    \left( 2 \frac{(L_0 + y) (L_1 + x) (R_0 + I_0 - y) (R_1 + I_1 - x)}{(L_0 + L_1 + x +y)^3 (R_0 + R_1 + I_0 + I_1 - x - y)^3} \right) =
\end{gather*}

\begin{equation*}
    \frac{\left( (L_0 + y) (R_1 + I_1 - x) - (L_1 + x) (R_0 + I_0 - y) \right)^2}{(L_0 + L_1 + x +y)^3 (R_0 + R_1 + I_0 + I_1 - x - y)^3}
\end{equation*}
\normalsize

Because of the square, the numerator is always non-negative. Since $x \leq I_1$ and $y \leq I_0$ the denominator is also always non-negative. Therefore the second principal minor is non-negative.

Since the first principal minor is non-positive and the second principal is non-negative, the hessian is negative semi-definite. Since the hessian is negative semi-definite the function is concave with respect to $x$ and $y$.

\subsection*{Finding the Maxima}
To find the maximum of the function $f(x, y)$ we first find all critical points by setting both the partial derivative with respect to $x$ and the partial derivative with respect to $y$ equal to zero:

\begin{equation*}
    \begin{cases}
        \frac{\partial}{\partial x} \left(\frac{(L_0 + y) (L_1 + x)}{L_0 + L_1 + x + y}
        +
        \frac{(R_0 + I_0 - y) (R_1 + I_1 - x)}{R_0 + R_1 + I_0 + I_1 - x - y}\right) = 0\\[6pt]
        \frac{\partial}{\partial y} \left(\frac{(L_0 + y) (L_1 + x)}{L_0 + L_1 + x + y}
        +
        \frac{(R_0 + I_0 - y) (R_1 + I_1 - x)}{R_0 + R_1 + I_0 + I_1 - x - y}\right) = 0
    \end{cases}
\end{equation*}


This system of equations has three solutions that we will each examine, only one of which lies in our region of interest ($x \in [0, I_1], y \in [0, I_0]$):
\begin{itemize}
    \item $x = - L_1$, $y = - L_0$: For any value of $L_0$ and $L_1$ other than $0$ the solution will fall outside of the ranges defined for $x$ and $y$.
    \item $x = R_1 + I_1$, $y = R_0 + I_0$: For any value of $R_0$ and $R_1$ other than $0$ the solution will fall outside of the ranges defined for $x$ and $y$.
    \item $x: \text{free}$, $y = \frac{L_1 R_0 - L_0 R_1 - L_0 I_1 + L_1 I_0 + (L_0 + R_0 + I_0) x}{L_1 + R_1 + I_1}$: These solutions lie on a line through our region of interest.
\end{itemize}

So now we have found a line that lies in our region of interest and consists of critical points:

\begin{equation*}
    y = \frac{L_1 R_0 - L_0 R_1 - L_0 I_1 + L_1 I_0 + (L_0 + R_0 + I_0) x}{L_1 + R_1 + I_1}
\end{equation*}

In the previous section we proved $f(x, y)$ was in fact concave with respect to $x$ and $y$. Therefore this line of critical points is a line of maxima of the function.

\subsection*{Integer Solutions}
Instead of a line we need a single point as a solution. In section 4 we mentioned that we select the point on the line that is closest to the starting values of $x$ and $y$ ($LI_0, LI_1$) but we did not explain how to compute this point. To efficiently compute the closest point on the line we can use the fact that it lies on a second line perpendicular to the first line. So we find the perpendicular line that passes through $(LI_0, LI_1)$ and compute its intersection with our first line. This intersection is $(x', y')$. We can first write the solution line we found in a more common format:


\small
\begin{equation*}
    \frac{L_0 + R_0 + I_0}{L_1 + R_1 + I_1} x - y + \frac{L_1 R_0 - L_0 R_1 - L_0 I_1 + L_1 I_0}{L_1 + R_1 + I_1} = 0
\end{equation*}
\normalsize

\noindent The point on this line closest to $(LI_0, LI_1)$ is then:

\scriptsize
\begin{multline*}
    x' = \frac{-\left(-LI_0 - \frac{L_0 + R_0 + I_0}{L_1 + R_1 + I_1} LI_1\right)}{\left( \frac{L_0 + R_0 + I_0}{L_1 + R_1 + I_1} \right)^2 + 1}\\
    - \frac{\frac{L_0 + R_0 + I_0}{L_1 + R_1 + I_1} \frac{L_1 R_0 - L_0 R_1 - L_0 I_1 + L_1 I_0}{L_1 + R_1 + I_1}}{\left( \frac{L_0 + R_0 + I_0}{L_1 + R_1 + I_1} \right)^2 + 1}
\end{multline*}

\begin{equation*}
    y' = \frac{\frac{L_0 + R_0 + I_0}{L_1 + R_1 + I_1} \left(LI_0 + \frac{L_0 + R_0 + I_0}{L_1 + R_1 + I_1} LI_1\right) + \frac{L_1 R_0 - L_0 R_1 - L_0 I_1 + L_1 I_0}{L_1 + R_1 + I_1}}{\left( \frac{L_0 + R_0 + I_0}{L_1 + R_1 + I_1} \right)^2 + 1}
\end{equation*}
\normalsize

Whenever we want to compute $S_{robust}$ we can now in constant time determine $x', y'$ and round them to the nearest integers. If $x'$ or $y'$ happen to lie outside of the regions $[0, I_1]$ and $[0, I_0]$ respectively, we can clip their values to the boundaries.

\end{document}